# AirRL: A Reinforcement Learning Approach to Urban Air Quality Inference

Huiqiang Zhong
BAIDU
China
zhonghuiqiang@baidu.com

Cunxiang Yin
BAIDU
China
yincunxiang@baidu.com

Xiaohui Wu
BAIDU
China
wuxiaohui02@baidu.com

Jinchang Luo
BAIDU
China
luojinchang@baidu.com

JiaWei He
BAIDU
China
hejiawei@baidu.com

## ABSTRACT

Urban air pollution has become a major environmental problem that threatens public health. It has become increasingly important to infer fine-grained urban air quality based on existing monitoring stations. One of the challenges is how to effectively select some relevant stations for air quality inference. In this paper, we propose a novel model based on reinforcement learning for urban air quality inference. The model consists of two modules: a station selector and an air quality regressor. The station selector dynamically selects the most relevant monitoring stations when inferring air quality. The air quality regressor takes in the selected stations and makes air quality inference with deep neural network. We conduct experiments on a real-world air quality dataset and our approach achieves the highest performance compared with several popular solutions, and the experiments show significant effectiveness of proposed model in tackling problems of air quality inference.

## CCS CONCEPTS

- ***

## 1 INTRODUCTION

There has been increasing urbanization in some developing countries, like China and India, in the past few decades. With urbanization, the air pollution is becoming a serious problem, posing a threat to public health. The relationship between health and exposure to air pollutants has been established in several studies [1-4]. To better control air pollution and assess health outcomes, it is necessary to get accurate measurements of air quality in both time and space. While the government has strived to build more air quality monitoring stations, current existing stations are still quite sparse, due to the exorbitant building costs. Consequently, it is of vital importance to infer the fine-grained air quality in urban areas without monitoring stations.

Currently, there are many methods to infer air quality in urban areas. Physically based methods [15] aim to infer air quality in areas without monitoring stations through simulating the propagation process of the air pollutants. These methods are based on the assumption that air pollution propagation conforms to probability distributions, for example Gaussian distribution. However, these assumptions do not always apply in real physical settings.

In recent years, air quality inference based on big data has become an important method among researchers. This method attempts to learn the impacts that environmental factors (e.g. meteorology, points of interest (POI), human activity, traffic, etc.) have on the air quality and the air quality relationship between areas with and without monitoring stations. The first problem to be addressed is how to establish the relationship between various environmental factors and monitoring data. A co-training framework is proposed by Zheng [2], which consists of two separated classifiers, a spatial classifier that takes spatially related features and a temporal classifier that involves temporally related features. Recently, deep learning models have been widely used to address this problem. Yi [9] proposed a deep neural network based approach that consists of a spatial transformation component and a deep distributed fusion network.

We primarily consider two kinds of relationships between areas with and without monitoring stations: their environmental similarity and the air pollution propagation relationship. Under stable weather conditions, the local air quality are largely determined by its surrounding environment. Thus, when inferring air quality in areas without monitoring stations under this condition, monitoring stations surrounded by similar environment are selected. Under other weather conditions, the local air quality is also subject to the air pollutants' propagation from other areas in addition to its local environment. Thus, it is necessary to select the monitoring stations from those areas. Therefore, another major problem to be addressed is how to dynamically select relevant monitoring stations to infer the air quality at target location under time-varying weather conditions. Zheng [2] selected air quality monitors randomly in their model to infer the unknown grids. Chen [5] adopted a K-Nearest Neighbor (KNN) strategy to select a subset of monitoring stations and only modeled the effects of the selected monitoring station data for inference. Unfortunately, the random selection strategy might cause the inconsistency problem, while the features from K Nearest stations are not necessarily the most effective, and the significance of the same stations may vary with time and meteorology. Cheng [3] combined feedforward and recurrent neural networks to model static and sequential features as well as an attention-based pooling layer to learn the weights of features from different monitoring stations automatically. The attention-based method integrated the data from all available stations during air quality inference. Nevertheless, we found that using parts of closely related monitoring stations for weight learning based on attention mechanism achieve better results than using all monitoring stations. In the physical world, weather is ever-changing and not all monitoring stations contribute to the inference of air quality for target location. For example, under circumstances where pollution is largely transported by wind, the monitoring stations in the upwind direction contribute to inferring air quality for target location, whereas stations in the downwind direction have few impacts. During air quality inference, if the monitoring station is located in a area from where it exerts no air pollution propagation towards the target location, using its data would worsen the inference results. It is critical to dynamically select important monitoring stations from all stations.

In this paper, we address the aforementioned problems by introducing a reinforcement learning model [6,7], named AirRL, for fine-grained air quality inference. AirRL consists of two modules: a station selector and an air quality regressor. Air quality regressor adopts deep neural networks to model heterogeneous data (air quality, POI, road networks, weather conditions), and learn the interactions between complex features. An original attempt of AirRL is to utilize reinforcement learning to select the important monitoring stations when inferring air quality in areas without monitoring stations. The contributions of our paper are as follows:

1) Based on reinforcement learning framework, we propose a novel model for air quality inference problem named AirRL. To our best knowledge, this paper is the first work to apply reinforcement learning to the task of air pollution inference.

2) We formulate station selection as a reinforcement learning problem, which enables our model to dynamically select optimal monitoring stations for air quality inference. Our model outperforms other approaches, which proves AirRL can distinguish the importance of monitoring stations.

## 2 OVERVIEW

### 2.1 Problem Formulation

In this paper, our goal is to infer the fine-grained urban air quality using the data of points of interest (POI), road net-works and meteorology of a city, air quality from the monitoring stations. An air quality index is used by government agencies to measure or forecast the air quality. Individual air quality index (IAQI) refers to the measurement of certain air pollutant in the monitoring station. We define the area where the air quality is to be inferred as the target location. We formulate the air quality inference as a regression problem. For a certain air pollutant, given the features in the area where the monitoring stations are located, IAQI $F_{as} = \{F_{as}^t\}_{t=1}^T$, POI data $F_{ps}$, road network $F_{rs}$ and meteorological data $F_{ms} = \{F_{ms}^t\}_{t=1}^T$, as well as the corresponding data in the area where the target location is located, POI data $F_{pl}$, road network $F_{rl}$ and meteorological data $F_{ml} = \{F_{ml}^t\}_{t=1}^T$, we aim to infer the IAQI in the target station. Since different air pollutants are monitored independently, we train a different model for every air pollutant. This paper focuses on the inference of PM2.5 and PM10.

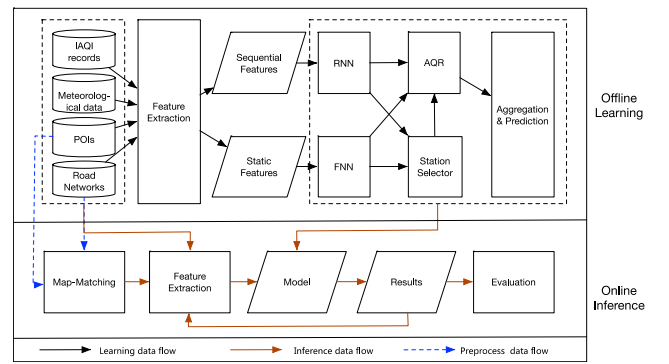

**Figure 1:** Framework of proposed system

### 2.2 Framework

Our proposed framework for air quality inference problem is shown in Figure 1. The framework consists of the major components: offline learning and online inference.

In offline learning, we build a deep learning regression model for IAQI. Temporally-related features (e.g., meteorology and IAQI) and spatially-related features (e.g., POI and road network) are used to construct training samples. Then the training samples are fed to a model based on reinforcement learning framework, which consists of two modules: a station selector and an air quality regressor. Station selector selects important monitoring stations (related to target location), and air quality regressor predicts the IAQI values based on selected stations. Following reinforcement learning policy, we take station selector as agent and use mean squared error (MSE) of the predicted value as environment reward. In order to take advantage of the deep learning approach, we build air quality regressor with neural networks, including feedforward neural network (FNN) and long short-term memory (LSTM) [8].

In online inference, the procedure tries to predict IAQI for target location in a given time period with trained model. Firstly, in feature extraction module, temporally-related feature is calculated online, but spatially-related features such as road network data are retrieved from database computed in offline process. Then station selector module will be triggered to create the relevant stations list, and the air quality regressor module predicts the final prediction value for each IAQI. Then we can evaluate the system with observed data.

## 3 METHODOLOGY

### 3.1 Feature Extraction

The features used in this paper have also been employed in previous works [2, 3], including meteorological features, POI features, road network features and monitoring features. We aim to predict air quality for target location $l$. The affecting region's radius $\bar{d}$ is set to 2 kilometers.

**Meteorological features $F_m$.** Air quality is strongly influenced by meteorology. For example, in general, the higher the wind speed, the more air pollutants are dispersed and the lower their concentration. We identify six features: temperature, humidity, barometer pressure, wind speed, wind direction and weather. Weather types and wind direction are categorical features with 17 and 10 categories each, and we represent them with one-hot encoding. Other features are numerical, and we adopt zero-mean normalization to normalize them. It's done through this formula: $x_* = (x - \mu)/\sigma$, where $\mu$ is the mean of observed data and $\sigma$ is the standard deviation.

**POI features $F_p$.** A point of interest (POI) is a specific location, such as factory, hotel which usually associates with a name, coordinates, category, and other attributes. POI has a great impact on air quality. For example, for typically congested factory neighborhood, its air quality tends to be bad while those surrounded by parks usually has better air quality. We consider 12 POI categories employed in [2, 3]. In fact, the upwind POI have greater influence on the air quality of target location $l$ while the downwind POI has less influence. The direction of POI relative to the target location $t$ is another feature, which is divided into four categories. The number of each POI category within a region is one feature, and POI's direction to target location as another. The POI features add up to 48 dimensions.

$$F_p = \{p_{d,c} \mid d \in D_p, c \in C_p, distance(p, l) < \bar{d}\}$$

Where $D_p$ is the four directions and $C_p$ is the set of POI categories, and $p_{d,c}$ is the number of POI under direction $d$ and category $c$.

**Road network features $F_r$.** A road network is a system of interconnecting lines and points in an area, made up of various types of road segments. A road segment may contain a list of points and two points connect by a directed edge. Actually, vehicles are known to be an important source of urban air pollutants [14]. The structure of a road network has a strong correlation with traffic pattern. We divide all road segments into four categories: highway, trunk, railway and others. To capture the intensiveness of road segments in different types, we measure the total length of road segments per category within a region as a feature.

$$F_r = \{length(r_c) \mid c \in C_r, r \in region(\bar{d})\}$$

where $C_r$ is the set of road categories, $r_c$ is the total length of road segments under category $c$.

**Monitoring features $F_{as}$ and $F_{ds}$.** The air quality features $F_{as}$ of monitoring station contain a sequence of observed IAQI values over time. In addition, we also extract the distance and direction feature $F_{ds}$ from monitoring station to target location.

For target location $l$, we extract features of POI $F_{pl}$, road network $F_{rl}$, and meteorological features $F_{ml}$. For each monitoring station $s$, we extract features $F_{ps}, F_{rs}, F_{ms}, F_{as}$ and $F_{ds}$ within the monitoring station affecting region.

### 3.2 Proposed Model

The proposed model is based on a reinforcement learning framework and consists of two components: the air quality regressor (AQR) and the station selector (SS). There exist dozens of monitoring stations, but not all monitoring stations contribute to inferring the air quality of target location. Some monitoring stations are too far away and some are in the downwind direction of the target location $l$, which may have noisy influence. POI features, meteorological features, road network features, and monitoring features jointly determine whether monitoring station contributes to the target location. The air quality regressor use features of target location and monitoring station to infer the air quality of target location. In the station selector, each monitoring station has a corresponding binary action $a_i$ to indicate whether or not the $i$-th station will be selected for air quality inference. The state $s_i$ is represented by the $i$-th candidate station representation, the average of chosen stations representation, and the target location representation. The station selector samples an action given the current state according to a stochastic policy. When training the selector, the station selector distills the monitoring stations to affect the output of the air quality regressor. Then, the air quality regressor gives reward to the station selector. The selector updates its parameters iteratively. Figure 2 shows the framework of proposed model.

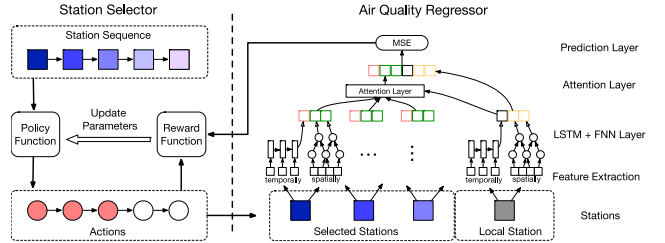

**Figure 2: Overall process of AirRL**

**Air Quality Regressor (AQR)**

**Regressor.** In the air quality regressor, we use a deep neural network architecture. There are totally four layers in the regressor model.

The first layer is feature extraction layer, which extracts the temporal features and spatial features for each station and target location.

$$x_s^S = (F_{ps} \oplus F_{rs} \oplus F_{ds}) \quad (3-1)$$
$$x_s^T = \{(F_{ms}^t \oplus F_{as}^t)\}_{t=1}^T \quad (3-2)$$
$$x_l^S = (F_{pl} \oplus F_{rl}) \quad (3-3)$$
$$x_l^T = \{F_{ml}\}_{t=1}^T \quad (3-4)$$

where $x_s^S$ is the spatial input feature of monitoring station, $x_s^T$ is the temporal features of monitoring station, $x_l^S$ is the spatial input feature of target location, $x_l^T$ is the temporal features of target location, and $\oplus$ is an operator of concatenation.

In the second layer, there are two unshared multilayer perceptron (MLP) to encode the spatial features from the target location and monitoring stations, respectively. Similarly, there are two unshared single direction LSTM to encode the temporal features, respectively.

$$h_*^S = W_*^S \cdot x_*^S + b_*^S \quad (3-5)$$
$$h_*^T = LSTM_*(x_*^T) \quad (3-6)$$

where $x_*^S$ is the spatial input feature, $x_*^T$ is the temporal input feature, $W_*^S$ and $b_*^S$ are the parameters of spatial MLP, $LSTM_*$ is temporal features encoder. Specially, $*$ represent the target location or monitoring stations, it could be $l$ or $s$.

Then, we use two unshared MLP to hybrid spatial representation and temporal representation. Specifically, for spatial features we choose POI features, road network for target location, and add distance feature for monitoring stations. For temporal features, we only use meteorological data as temporal features for target location, but also use air quality data for monitoring station.

$$z_* = W_*^h \cdot (h_*^S \oplus h_*^T) + b_*^h \quad (3-7)$$

where $z_*$ is the spatial-temporal representation, $W_*^h$ and $b_*^h$ are the parameters of hybrid MLP, and $\oplus$ is an operator of concatenation. Thus, we get a location representation and multi-station representation. The monitoring stations are selected by the station selector, which will be described in later section.

e use attention mechanism to calculate the weights of selected stations

In the third layer, we use attention mechanism to calculate the weights of selected stations, which can be written as:

$$a'_{s_i} = w_{s_i}^{a_2} \phi(W^{a_1}(z_l \oplus z_{s_i}) + b_{s_i}^{a_1}) + b_{s_i}^{a_2} \quad (3-8)$$

$$a_{s_i} = \frac{\exp(a'_{s_i})}{\sum_k \exp(a'_{s_k})} \quad (3-9)$$

$$z_s = \sum_i^n a_{s_i} \odot z_{s_i} \quad (3-10)$$

where $W^{a_1}$, $W_{s_i}^{a_2}$, $b_{s_i}^{a_1}$ and $b_{s_i}^{a_2}$ are the parameters of attention mechanism. $a_{s_i}$ is the attention weight scalar for each selected stations. $z_s$ is the weighted sum of $z_{s_i}$, which represents all of the selected stations. $\odot$ means element wise multiplication.

In the fourth layer, we concatenate the location representation and the station representation, and feed it into a MLP layer to output the air quality value.

$$y = W^o \cdot [z_l \oplus z_s] + b^o \quad (3-11)$$

where $z_l$ is the target location representation, $(z_{s_1}, \dots, z_{s_n})$ is all the n selected station representations and $y$ is the output value. $W^o$ and $b^o$ is the parameters of output layer.

**Loss function.** The objective of the air quality regressor is to make the predicted result close to true value as much as possible. So we define the loss function of the air quality regressor using mean squared error as follows:

$$MSE = (y - \hat{y})^2 \quad (3-12)$$

where $y$ is the predicted value, and $\hat{y}$ is the ground truth value.

**Station Selector (SS)**

Since not all monitoring data contributes to predicting air quality in the target location. We propose to leverage the reinforcement learning to our AirRL model that choose the relevant stations automatically. Existing methods for station selection include random selection strategy [2], k-nearest neighbor strategy [5] and attention mechanism [3]. Unfortunately, the random selection strategy might cause the inconsistency problem, while the features from K Nearest stations are not necessarily the most effective, and the significance of the same stations may vary with time and meteorology. In addition, we found that using parts of closely related monitoring stations for weight learning based on attention mechanism achieve better results than using all monitoring stations. The major component of our model is to select relevant monitoring stations using reinforcement learning to improve the inference results in target locations prior to attention layer, and we call this component the station selector.

The station selector is the agent, which interacts with the environment. For each target location, we feed all monitoring stations to station selector, which will decide which monitoring stations contribute more to the inference. Thus, the station selector abandons stations with limited influence, such as the station that is far from target location. The station selector uses a network to choose monitoring stations based on features of target location, current station and already selected stations.

We treat candidate stations in a training sample as a sequence, $Z = \{z_{s_1}, \dots, z_{s_n}\}$ and compute a reward until all selection is finished in a sequence. We define the action as selecting a station or not. The reward is computed once all the selection decisions $\{a_1, \dots, a_n\}$ are completed on one sequence. The station selector selects stations among all monitoring stations. The air quality regressor then makes air quality inference using the selected stations. The state, action and reward are described in detail as follows. The description here is based on only one sequence.

**State.** The state $s_i$ includes target location representation $z_l$, the current candidate station representation $z_{s_i}$, and the average of selected stations representation.

1) The target location representation $z_l$ is from target location hybrid MLP layer which combines the POI, road network and meteorology.

2) The current candidate station representation $z_{s_i}$ is from station hybrid MLP layer which combines the POI, road network, meteorology, air quality, and distance-direction relative to target location.

3) The selected station representation is from the average of already selected stations representation.

**Action.** We define an action $a_i \in \{0,1\}$ to indicate whether the station selector will select the i-th station of the sequence $Z$ or not. We sample the value of $a_i$ by its policy function $\pi_\theta(s_i, a_i)$, where $\theta$ is the parameters to be learned. In this work, we adopt a logistic function as the policy function:

$$\begin{aligned}\pi_\theta(s_i, a_i) &= p(a_i|s_i) \\ &= a_i * \delta(W * F(s_i) + b) \\ &\quad + (1 - a_i) * (1 - \delta(W * F(s_i) + b))\end{aligned} \quad (3-13)$$

where $F(s_i)$ is the state feature vector, and $\delta$ is the sigmoid function with the parameter $\theta = \{W, b\}$.

**Algorithm 1** AirRL Trainning Procedure
1. Initialize the parameters of the AQR model and the policy network of station selector with weights respectively by sampling from the uniform distribution between −1 and 1.
2. Pre-train the AQR model to regress the air quality by minimizing mean squared error.
3. Pre-train the policy network by running Algorithm 2 with the AQR model fixed.
4. Run Algorithm 2 to jointly train the AQR model and the policy network until convergence.

---

**Algorithm 2** Reinforcement Learning Algorithm for the Station Selector
**Input:** Episode number $L$. Training data
  $X = \{X^1, X^2, ..., X^N\}$. A AQR and a policy
  network model parameterized by $\Phi$ and $\Theta$,
  respectively
Initialize the target networks as: $\Phi' = \Phi$, $\Theta' = \Theta$
1: **for** $episode\ l = 1$ to L **do**
2:    Shuffle **X**
3:    **for** $X^k \in \mathbf{X}$ **do**
4:       Sample selection actions for each monitoring
5:       station in $X^k$ with $\Theta'$:
6:       (To be clear, we omit the superscript k below)
7:       $A = \{a_1, ..., a_{|X|}\}, a_i \sim \pi_{\Theta'(S_i, a_i)}(X)$
8:       Compute delayed reward $r(s_{|X|+1}|X)$
9:       Update the parameters $\Theta$ of station selector:
10:      $\Theta \leftarrow \Theta + \alpha \sum v_i \nabla_\Theta \log \pi_{\Theta(s_i, a_i)}$, where
11:      $v_i = r(s_{|X|+1}|X)$
12:    **end**
13:
14:    Update $\Phi$ in the AQR model
15:    Update the weights of the target networks:
16:      $\Theta' = \tau\Theta + (1-\tau)\Theta'$
17:      $\Phi' = \tau\Phi + (1-\tau)\Phi'$
18: **end**

**Reward.** The reward function is an indicator of the utility of the chosen stations. For a sequence $Z = \{z_{s_1}, ..., z_{s_n}\}$, we sample an action for each station, to determine whether the current station should be selected or not. We assume that the model has a terminal reward when it finishes all the selection. The reward is defined as follows:

$$r(s_i|\hat{Z}) = \begin{cases} 0 & i < |Z| \\ e^{-(y-\hat{y})^2} & i = |Z| \end{cases} \quad (3-14)$$

where $\hat{Z}$ is the set of selected stations, $y$ is the predicted value given $\hat{Z}$, and $\hat{y}$ is the ground truth value. The above reward evaluates the overall utility of all the actions made by the policy. The objective function of the station selector is consistent with the air quality regressor, and it motivates the station selector to minimize the loss of air quality regressor.

**Optimization.** For a sequence $Z$, the objective function is defined as follows to maximize the expected reward.

$$J(\theta) = E[\sum_i r(s_i|\hat{Z})] \quad (3-15)$$

where $J_\theta$ is the value function, represents the expected reward that we can obtain by starting at certain state $s_0$. For each sequence $Z$, we sample an action for each state sequentially according to the current policy. We then get a sampled action sequence $\{a_1, ..., a_n\}$ and a corresponding terminal reward. The current policy is updated using the following gradient:

$$\theta \leftarrow \theta + \beta \sum_i r(s_{|B|}|\hat{Z}) \nabla_\theta \log(\pi_\theta(s_i, a_i)) \quad (3-16)$$

where $\beta$ is learning rate.

**Model Training**
As the station selector and the air quality regressor are correlated, we train them jointly. The joint training process is described in Algorithm 1. We first pretrain the AQR with all monitoring stations, and then pretrain the policy function by computing the reward with the pretrained AQR, while the parameters of the AQR model are frozen. Then, the station selector and the air quality regressor will be jointly trained in Algorithm 2. The air quality regressor provides a mechanism for computing the rewards of the selected sentences to refine the station selector. In order to have a stable update, we take advantage of a target policy network and a target AQR network, respectively.

## 4 EXPERIMENT

To evaluate the performance of our proposed model, we compare it against other methods like interpolation method, KNN and Attention Neural Network. For better comparison, we implement all the mentioned methods and evaluate them on the same dataset.

**Table 1: Input feature setting**

| Data | Feature | Detail | Dim |
|---|---|---|---|
| Target Location | POI | 4(directions) × 12(categories) | 48 |
| | Road Network | 4(categories) | 4 |
| | Meteorology | 17 (weather) 10 (wind direction) 4(other features) | 31×24 |
| Monitoring stations | POI | 4(directions) ×12(categories) | 48 |
| | Road Network | 4(categories) | 4 |
| | Meteorology | 17 (weather) 10 (wind direction) 4(other features) | 31×24 |
| | Air Quality | 1(IAQI) | 1×24 |
| | Distance And Direction | 1(distance) 8 (direction) | 9 |

### 4.1 Dataset

To evaluate the performance of AirRL, we use the Microsoft dataset collected in Beijing, China, similar to the dataset in [2][3]. The air quality data [13] are recorded at 36 air quality stations every hour, from 2014/05/01 to 2015/04/30. We only select 30 stations for experiment because remaining stations have too much missing value. We focus on PM2.5 and PM10, which are converted to IAQI metrics following the Chinese AQI standard. Meteorological data are collected at a district (or city) level. We locate the monitored station's district, and use its meteorological data as the station data. POI data are collected from Map World APIs [12] and 12 categories of POI are considered. The road network data are collected from OpenStreetMap [11] and four categories are considered: highway, trunk, railway and others. In our experiment, air quality and meteorological data of the last seven hours are considered.

We split the observed dataset into the training and test sets with the ratio of 2 to 1, based on stations. For early stopping, 10% of training data are selected as the validation set. The features of target location include POI, road network and meteorology. In addition to the aforementioned features, monitoring stations also include features of air quality and distance-direction. The details of input features are listed in Table 1. Categorical features are represented as one-hot encoding, and numerical features are transformed with zero-mean normalization. Meteorology and air quality are both temporal features, which have shown apparent periodic fluctuation on the daily basis. Meanwhile, experiments have shown that using a window of 24 hours could achieve better results.

## 4.2 Settings and Compared Methods

Firstly, we describe our setting for AirRL, and then the compared methods are listed.

1) In the Air Quality Regressor, we use different setting for the features of target location and monitoring station. For spatially-related features, one basic FNN layer (L=1) with 25 neurons is constructed for target location, and one basic FNN layer (L=1) with 50 neurons for monitoring stations. For temporally-related features, we build two LSTM layers with 150 memory cells per layer and use two individual LSTM for target location and monitoring station. Finally, in the predicting layer, we build one layer of high-level FNN network (L=1) with 50 neurons to combine target location features and monitoring stations features.

2) In the Station Selector, we construct one basic FNN layer (L=1) with 16 neurons.

In all layers, we use RELU as the activation function and dropout to prevent overfitting. We initialize all the parameters with a uniform distribution.

We compare AirRL with the following baselines:

1) Linear interpolation (Linear): This method [10] adopts distance-weighted value as interpolation parameter.

$$\text{IAQI} = \sum_i \frac{IAQI_i * (1/d_i)}{\sum(1/d_i)} \quad (4-1)$$

where $d_i$ is the distance from target location to the $i$-th monitoring station, and $IAQI_i$ is the $i$-th observed value.

2) Gaussian Interpolation (Gaussian): This interpolation method is based on a Gaussian distribution.

$$\text{IAQI} = \sum_i IAQI_i \times \frac{1}{\sigma\sqrt{2\pi}} e^{-\frac{d_i^2}{\sigma^2}} \quad (4-2)$$

where $\sigma$ represents the average distance between two monitoring stations.

3) K Nearest Neighbors (KNN): This method selects the K nearest monitoring stations and compute the average IAQI value. We set K to be 5 in our experiments.

4) ADAIN: This model [3] combines feedforward and recurrent neural networks to model static and sequential features as well as an attention-based pooling layer to learn the weights of features from different monitoring stations automatically.

5) Air Quality Regressor: This method is one part of AirRL, but without the station selector.

**Metrics** Firstly, the root mean squared error (RMSE) is employed to measure the performance of various regression approaches:

$$RMSE = \sqrt{\frac{\sum_i^N (y - \hat{y})^2}{N}} \quad (4-3)$$

Where N is the number of test dataset, $y$ is the ground truth value of IAQI, and $\hat{y}$ is the predicted value of IAQI.

We also measure the performance with classification metrics. Prior to evaluation, the regression values are converted into the corresponding IAQI levels. Accuracy is defined as:

$$Accuracy = \frac{|\{x \mid l(y(x)) = l(\hat{y}(x))\}|}{N} \quad (4-4)$$

Where $l(\cdot)$ is the function to convert continuous IAQI into discrete IAQI, consistent with the definition in Zheng [2]. $l(y(x))$ is the level of true value of IAQI and $l(\hat{y}(x))$ is the level corresponding to the predicted value of IAQI.

**Table 2: RMSE Comparison for station selection**

|  | PM2.5 | PM10 |
|---|---|---|
| AQR-18 | 39.0600 | 46.2411 |
| AQR-10 | 38.9074 | 45.5416 |
| AQR-5 | 38.1233 | 43.5950 |
| AQR-3 | 38.8807 | 44.5883 |
| AirRL | **35.7555** | **40.8122** |

## 4.3 Results

First, we verified the better inference results achieved by using parts of closely related monitoring stations based on attention mechanism compared against using all of the monitoring stations. The distances between monitoring stations and target station are calculated, and the nearest 18, 10, 5, and 3 monitoring stations to the target station are selected as 4 scenarios to compare the AQR model results, represented as AQR-18, AQR-10, AQR-5 and AQR-3, respectively. In addition, the AQR models for the 4 scenarios are also compared against the proposed AirRL model. The RMSE results are shown in Table 2, which suggests that AQR-5 achieves better results than the other 3 scenarios by selecting the nearest 5 monitoring stations. While attention mechanism introduces the ability to learn the weights of different monitoring stations on the target station, even better results could be achieved by selecting some closely related monitoring stations prior to the attention layer. Thus, it is an essential step to select relevant monitoring stations. The selection of the nearest stations is not the most effective strategy, and the experiment results indicate that dynamically selecting monitoring stations can achieve the best results due to the fact that AirRL has incorporated multiple factors including environmental similarity, distance and meteorology based on the reinforcement learning framework.

We also compare the AirRL model with other mainstream methods, the RMSE results of which are shown in Table 3. AirRL obtains the lowest RMSE for predicting all air pollutants. ADAIN achieve the second best performance, which uses more features and perform well regarding feature interaction. LSTM handles temporal features effectively and FNN captures spatial features well. Besides, ADAIN adopts the attention mechanism and gets slightly better

AirRL: A Reinforcement Learning Approach to Urban Air Quality Inference

performance. Linear, Gaussian and KNN get worse performance in that they use very few features and only consider the distance from the monitoring stations to the target location. The RMSE of AirRL is lower than ADAIN with 3.1 absolute points for PM2.5 inference, and 2.9 points for PM10 inference because AirRL has an additional station selector. This indicates that the station selector can select associated monitoring stations and filter out useless stations. The reason why AirRL is better than ADAIN may be that the attention mechanism aims to learn the weights of the monitoring stations, and the selector aims to choose associated monitoring stations. Some noisy monitoring stations might lead to inaccuracy while inferring air quality for target location. Abandoning those irrelevant stations, such as stations too far away, can improve the capability of the model.

In addition, we convert the regression results into the corresponding IAQI levels and compare AirRL with other methods using classification performance metrics. The results are shown in Table 4. The AirRL also obtains the best result.

**Table 3: RMSE comparison of different methods**

|          | PM2.5   | PM10    |
|----------|---------|---------|
| Linear   | 39.1895 | 47.3336 |
| Gaussian | 39.4427 | 47.3372 |
| KNN      | 40.5613 | 47.8476 |
| ADAIN    | 38.8379 | 43.7186 |
| AirRL    | **35.7555** | **40.8122** |

**Table 4: Accuracy comparison of different methods**

|          | PM2.5  | PM10   |
|----------|--------|--------|
| Linear   | 0.6389 | 0.5924 |
| Gaussian | 0.6343 | 0.5876 |
| KNN      | 0.6219 | 0.5792 |
| ADAIN    | 0.6813 | 0.6151 |
| AirRL    | **0.7137** | **0.6499** |

To better understand the effects of AirRL, we visualize the selected monitoring stations in Figure 3. Two target locations (location 1 and location 2) under two meteorological conditions (time 1, time2) are considered in the study. Figure 3(a) and 3(c) show that AirRL selects nearby monitoring stations to infer air quality for both target location 1 and 2 in calm weather at time 1. However, at time 2 in windy weather, the monitoring stations in the upwind direction have a greater impact on target locations. AirRL selects northwest monitoring stations for target location 1 in Figure 3(b). In addition to the monitoring stations in the upwind direction, AirRL also selects other monitoring stations for target location 1 in Figure 3(d). The reason may be that they have similar features of POI and road network. This shows that AirRL is capable of dynamically selecting important station data for prediction. Distance and meteorology are critical factors that determine the importance of selected stations.

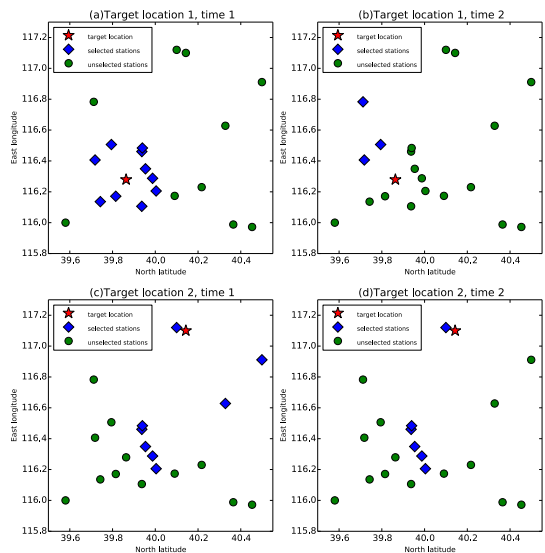

**Figure 3:** Selector visualization

## 5 CONCLUSIONS

In this paper, we presented a novel model for urban air quality inference based on a reinforcement learning framework. The model is composed of two modules: station selector to select relevant stations and air quality regressor to infer fine-grained air quality. We conduct empirical evaluation on several popular models, and the results show that our approach can achieve high accuracy in air quality inference.

Compared with existing approaches of air quality inference, our model is also able to dynamically select optimal air quality stations for inference in target locations by incorporating various model features.

To the best of our knowledge, this paper is the first work to apply reinforcement learning to the task of air pollution inference. In the future, we would like to apply our approach to other topics of air pollution problems, such as air pollution prediction.